# XMACNet: An Explainable Lightweight Attention based CNN with Multi-Modal Fusion for Chili Disease Classification

[1]Mr.Tapon Kumer Ray, [2]Dr. Rajkumar Y, [3]Dr. Shalini R, [4]Ms. Srigayathri K, [5]Dr. Jayashree S, [6]Ms Lokeswari P

[1]Student, Vellore Institute of Technology, Amaravati, Andhra Pradesh, India.
[2]Assistant Professor, Vellore Institute of Technology, Amaravati, Andhra Pradesh, India.
[3]Assistant Professor, Vellore Institute of Technology, Amaravati, Andhra Pradesh, India.
[4]Student, Angel College of Engineering and Technology, Tiruppur, Tamilnadu, India
[5]Assistant Professor, School of Computing, Department of Computer Science, Vel Tech Rangarajan Dr.Sagunthala R&D Institute of Science and Technology, Avadi, Chennai, India
[6]Assistant Professor, Department of Artificial Intelligence and Machine Learning, Bannari Amman Institute of Technology, Sathyamangalam, Erode.

Corresponding Author[2]: Dr. Rajkumar Y Email ID[2]: rajkumar.y@vitap.ac.in

**Abstract:** Plant disease classification via imaging is a critical task in precision agriculture. We propose XMACNet, a novel light-weight Convolutional Neural Network (CNN) that integrates self-attention and multi-modal fusion of visible imagery and vegetation indices for chili disease detection. XMACNet uses an EfficientNetV2S backbone enhanced by a self-attention module and a fusion branch that processes both RGB images and computed vegetation index maps (NDVI, NPCI, MCARI). We curated a new dataset of 12,000 chili leaf images across six classes (five disease types plus healthy), augmented synthetically via StyleGAN to mitigate data scarcity. Trained on this dataset, XMACNet achieves high accuracy, F1-score, and AUC, outperforming baseline models such as ResNet-50, MobileNetV2, and a Swin Transformer variant. Crucially, XMACNet is explainable: we use Grad-CAM++ and SHAP to visualize and quantify the model's focus on disease features. The model's compact size and fast inference make it suitable for edge deployment in real-world farming scenarios.

## 1. Introduction

Automated plant disease identification is essential for timely intervention and higher crop yields. Visual symptoms on leaves and fruits often precede severe crop losses, motivating the use of AI-based inspection [2, 7]. Convolutional neural networks (CNNs) have become the dominant approach for leaf disease classification due to their ability to learn complex visual features [1, 8]. However, classical CNNs can overfit when training data are limited and are often viewed as "black boxes". Furthermore, they typically rely on RGB imagery alone, potentially ignoring valuable multi-spectral information. This has been serving as a motivation for us to carry out the research work.

### 1.2 Major Contributions

To address these gaps, we develop XMACNet - an Xplainable MAss-sensing Convolutional Network. XMACNet is built on EfficientNetV2-S, a modern CNN backbone known for high accuracy with a small footprint [1]. We augment it with a self-attention module to capture global contextual cues and a multimodal fusion branch that integrates RGB inputs with vegetation indices (NDVI, NPCI, MCARI) computed from NIR/visible channels [2, 3]. The fusion of these modalities helps the model highlight plant stress signals that may be invisible in RGB space alone. We also emphasize interpretability: leveraging Grad-CAM++ [5] and SHAP [6], we provide visual and quantitative explanations of XMACNet's predictions. The key contributions include:

- ➢ **XMACNet Architecture:** A novel CNN combining EfficientNetV2-S with self-attention and a multimodal fusion scheme.
- ➢ **Vegetation Indices Fusion:** Integration of NDVI, NPCI, and MCARI alongside RGB inputs, exploiting chlorophyll and pigment cues that are indicative of diseases [2, 3].
- ➢ **Dataset and Augmentation:** A new chili disease dataset (12,000 images, 6 classes) augmented using StyleGAN2-based synthesis [4] to ensure data diversity and balance.
- ➢ **Explainability:** Use of Grad-CAM++ and SHAP to interpret model decisions, improving transparency in plant disease classification.
- ➢ **Lightweight Edge Deployment:** A compact, high-performance model suitable for deployment on resource-constrained devices, achieving faster inference and requiring fewer parameters than larger architectures [1, 7].

Collectively, these advances yield a chili disease classifier that is accurate, robust, interpretable, and practical for real-world use.

### 1.3 Organization of the Article

The proposed work has been organized as follows: Section 1 describes the introduction along with the motivation and the major contributions; section 2 involves the related works section along with the comparative analysis of various chilli disease classification methods; section 3 provides the proposed methodology; section 4 provides the experimental setup and the performance metrics used; section 5 involves the performance analysis; section 6 provides the comparative analysis of the proposed methodology to that of the existing benchmarks and section 7 provides the conclusion. Table.1 provides the chronicle of abbreviations of the proposed work.

**Table 1.** Chronicle of Abbreviations

| Abbreviation | Description |
|---|---|
| CNN | Convolutional Neural Network |
| XMACNet | Explainable Lightweight Attention based Convolutional Neural Network |
| NDVI | Normalized Difference Vegetation Index |
| NPCI | National Payments Corporation of India |
| MCARI | Modified Chlorophyll Absorption Ratio Index |
| RGB | Red Green Blue |
| ResNet | Residual Neural Networks |
| MobileNet | Mobile Neural Networks |
| GradCAM | Gradient Weighted Class Activation Mapping |
| SHAP | SHapley Additive exPlanations |
| GAN | Generative Adversarial Networks |
| AUC | Area Under the Curve |
| ROC | Receiver Operating Characteristic |
| XAI | Explainable Artificial Intelligence |
| DCGAN | Deep Convolutional Generative Adversarial Network |
| ES | Ensemble Stack |
| SVM | Support Vector Machine |
| RF | Random Forest |
| XGB | Extreme Gradient Boosting |
| PSO | Particle Swarm Optimization |
| LIME | Local Interpretable Model-agnostic Explanations |
| YOLO | You only Look Once |
| MHA | Multi-Head Attention |
| t-SNE | t-distributed Stochastic Neighbor Embedding |

## 2. Related Works

From the table.2 it is evident that Deep CNNs have revolutionized crop disease diagnostics. Prior works have applied ResNet, VGG, and Inception models to leaf imagery with good accuracy. For example, Albahli *et al.* developed *AgriFusionNet*, a lightweight model based on EfficientNetV2-B4 that fuses RGB and multispectral (drone) imagery. Combining image data with additional modalities (e.g. multispectral bands, environmental sensors) has been explored recently. Limited labeled data is a common challenge. Generative adversarial networks (GANs) have been used to synthesize plant disease images. The black-box nature of CNNs limits trust in agricultural contexts. Methods like Grad-CAM++5 generate saliency maps that localize image regions influencing decisions, and SHAP assigns feature importance values based on Shapley theory [33]. These techniques have been applied to interpret plant disease models, showing that correct predictions often focus on lesion areas. Thus the proposed work incorporates both methods to analyse XMACNet: Grad-CAM++ for visual heatmaps on leaves, and SHAP for pixel-level or channel-level contributions. This layered explanation helps validate that XMACNet attends to biologically relevant features (e.g. discoloration, spots) when classifying diseases.

## Table 2. Comparative Analysis of Research Works on Chilli Disease Classification

| Ref No | Problem Addressed | Learning Model | Datasets Utilized | Benefits | Limitations |
|---|---|---|---|---|---|
| [11] | Red Chilli Adulteration (with brick powder, even minuscule amounts) | Meta Classifier- ES (SVM, RF, XGBoost with Linear Regression, Data augmentation-DCGAN; Feature Selection-Particle Swarm Optimization (PSO); Explainable AI (XAI) - SHAP and LIME for transparency and interpretability. | Custom Laboratory Controlled Dataset – 250 Natural Samples (5 categories of adulteration). Augmented with 1000 synthetic samples (200 per category) generated by DCGAN. Handcrafted features (geometrical, texture, color, shape) extracted. | 1. High Accuracy-92.42% on synthetic Data; 2. Enhanced Interpretability and transparency; 3. Robustness: Effective in detecting micro-level adulteration (even at 1%). 4. Use of DCGAN addresses Data Scarcity 5. Overcomes Premature Convergence | 1. Adaptability to other adulterants or food products needs further exploration. 2. Relies on a laboratory- controlled dataset, which may not fully represent real- world variations. 3. Methods to reduce computational complexity for real-time or edge computing applications need exploration. |
| [12] | Rice & Apple Crop Disease Detection (irregular lesion shapes/sizes, traditional DL limitations), limitations of conventional Multi-Head Attention (MHA) | Enhanced Vision Transformer (ViT) network with an improved triplet Multi-Head Attention (t-MHA) function; Explainability study using LIME and t-SNE algorithms. | RiceApp Dataset, PlantVillage dataset-Publicly available (55K images, 38 classes). Both datasets augmented and pre-processed. | 1. High Accuracy-97.99% on RiceApp Dataset; 98.57% on PlantVillage Datasets | Some datapoints in t-SNE visualization still overlap due to similar symptomatic behaviour. |
| [13] | Potato Plant Disease Detection (from leaf images, addressing DL "black box" nature and computational expense) | Uses Multiple ML classifiers - LR, SVM, RF, GB, AB, NB; Uses Hough Transform (HT) and Discrete Wavelet Transform (DWT) for feature extraction across various color spaces. Explainable AI (XAI) promoted through feature importance analysis. | PlantVillage dataset from Kaggle. Contains healthy, early blight, and late blight images. Data augmentation applied to address initial class imbalance. | 1. Highest accuracy: LR achieved 0.99 in YCbCr color space, outperforming other models and SOTA techniques. 2. Feature importance analysis: DWT features significantly contributed to performance. 3. Low computational cost, suitable for precision agriculture. 4. Enhanced precision in disease identification. | 1. Still scope for improvement in using low computationally expensive feature extraction techniques. 2. Lack of proper datasets for many other plant diseases. |
| [14] | General Plant Disease Detection (based on visual features). | DenseNet169 | PlantVillage Dataset (54k images). Images used directly without noise reduction. | 1. High accuracy: Achieved 97.8% overall accuracy. 2. Efficient feature reuse and reduced computational cost due to its architecture. 3. Prevents overfitting using global average pooling and early stopping. | Training can be more time- consuming due to a larger number of layers. |
| [15] | Chilli Plant Disease (detection using vegetation indices). | 1. Support Vector Machine (SVM) 2. Classifier. Uses vegetation indices like NPCI, MCARI, NDVI, TCARI, and OSAVI as features. | 1. A custom database of normal and diseased chilli leaf samples, including spectral signatures. 2. Specific dataset size not detailed, but samples for "Forty Samples" of normal and diseased indices are shown. | 1. Effective classification: SVM with NPCI index showed a low 0.1 miss classification rate. 2. Identifies infected leaves using vegetation indices. | 1. Miss classification rates are relatively high for some indices (e.g., NDVI 0.4). 2. Limited dataset size details. |
| [16] | Chilli Plant Disease (distinguishing healthy, | Support Vector Machine (SVM) classifiers (Quadratic SVM and Medium Gaussian SVM highlighted). | Small custom dataset: 20 chilli plant images for training (100 segmented | 1. Correctly distinguishes healthy, diseased, and background areas with high accuracy (90.9% for background and healthy). | 1. Low accuracy (57.1%) for classification of cucumber mosaic. 2. Uses a very small dataset. |
| | unhealthy, and background parts of leaves). | Uses K- means clustering for image segmentation. | results) and 8 images for testing (40 segmented results). | 2. Quadratic SVM and Medium Gaussian SVM showed 90% training accuracy.<br>3. K-means clustering with 5 clusters provided optimum segmentation. | |
| [17] | Chilli Leaf Disease Identification (analyzing overall percentage of affected leaf and surrounding region, addressing manual cropping limitations). | YOLO; Uses K- means and Otsu for segmentation, and feature extraction based on shape and texture. | Images of chilli plant leaves with bounding box annotations for multiple leaves. Split into train and test data. | 1. Analyzes the overall percentage of affected leaf.<br>2. Uses image processing techniques like histogram equalization and brightness variations.<br>3. Supports detection of multiple leaves in a single image via bounding boxes. | Low overall mAP for test dataset (41.43%) and very low mAP for "not infected" class (17.60%) on test set. |
| [18] | Weligama Coconut Leaf Wilt Disease (WCLWD) and Coconut Caterpillar Infestation (CCI) (early diagnosis and severity assessment). | CNNs (ResNet50, DenseNet121, InceptionResNetV2, etc.) for WCLWD classification and severity assessment. Mask R-CNN (ResNet101 backbone, FPN) for CCI identification and classification. YOLOv5, YOLOv8, and YOLO11 for counting caterpillars. | Custom dataset (published on Kaggle) of coconut images from Sri Lanka. WCLWD: 9,258 images (classification), 3,307 (severity).<br>CCI: 1,600 images (classification), 1,400 (counting). Images preprocessed and augmented. | 1. High accuracy: CNNs yielded 90-97% for WCLWD; InceptionResNetV2 achieved 97% for severity. Mask R- CNN had 95.26% mAP for CCI. YOLOv5 was most effective for caterpillar counting.<br>2. Automated caterpillar counting minimized human inaccuracies. | 1. Model performance may be affected by extreme environmental conditions not fully represented (e.g., heavy occlusion, variable lighting).<br>2. Generalizability to other agricultural regions needs further validation.<br>3. Computational nature can be problematic for low-data or distant regions. |
| [19] | Corn Leaves Multi-disease Detection (Northern corn leaf blight, pest damage, brown spot, downy mildew; addressing feature variability, complex backgrounds, and efficiency). | GAF-Net (Gsconv- Adaptive Feature Fusion Network) based on YOLOv8. Incorporates Feature-Adaptive Spatial Feature Fusion (FASFF) and Grouped Spatial Convolution (GSConv) modules. Uses YOLOv8 backbone network with SPPF module. | Custom dataset of corn leaf images (1415 collected, 1219 selected) encompassing four major pest/disease types. Images standardized. | 1. Superior accuracy and robustness: Achieved mAP@50 of 80.8% (+4.1% over baseline) and mAP@50-95 of 52.6% (+3.3%).<br>2. Outperformed other SOTA models.<br>3. Effective handling of complex features.<br>4. Good balance between model efficiency and precision.<br>5. Overcomes data scarcity by constructing a comprehensive dataset.<br>6. Near-perfect detection accuracy (zero omissions) for small pests and in complex backgrounds. | 1. Expansion beyond current four disease types is needed.<br>2. High computational resource consumption (due to BERT model).<br>3. Does not involve knowledge reasoning.<br>4. Needs lightweight model variants for edge devices. |
| [20] | Chilli and Onion Leaf Diseases (multi-class classification; addressing intra-class similarity, complex backgrounds, and real-world variations). | Transformer-based deep learning framework evaluating MaxViT, Swin Transformer, Hornet, and EfficientFormer. Integrates Grad-CAM for visual explanations (XAI). | COLD dataset (custom curated) of 13,989 high- resolution images (chili and onion leaves) from agricultural environments in Karnataka, India. Covers nine disease classes. Extensive preprocessing applied. | 1. Highest accuracy: MaxViT achieved 95.75% on onion and 90.86% on chili dataset.<br>2. Enhanced model generalization. Real-time web application developed for practical field use.<br>3. Transparency and trust: Grad-CAM enhances interpretability for non-experts. | 1. Reduced performance on visually overlapping chili diseases.<br>2. High computational cost of transformer models.<br>3. Limited by the scope of disease types.<br>4. Future work needed for dataset expansion and model optimization for edge deployment. |
| [21] | Fruit Tree Leaf Diseases (classification of common diseases). | ResNet18 | Plant Village project dataset (54,305 RGB images across 38 classes). Split into 80% training, 20% evaluating. | 1. Commendable recognition rates (95.1% accuracy).<br>2. Suitable for real- world applications where data may be scarce. | 1. Generalizability may be constrained by the specific dataset.<br>2. Relatively small number of classes (38) compared to actual variety. |

| | | | | | 3. Relies exclusively on images; supplementary data could help. 4. Environmental factors (lighting) may affect performance; normalization needed. 6. Primarily focused on classification rather than severity. 7. Hyperparameter tuning was not implemented. |
|---|---|---|---|---|---|
| [22] | Paddy Crop Brown Spot Disease Detection (early-stage detection to reduce pesticide use). | Modified YOLOv9 model. Uses GELAN for feature extraction, PGI for training efficiency, and NMS for post-processing. | Custom brown spot paddy images dataset, collected and labeled with RoboFlow. | 1. High accuracy: Achieved 94.9% precision, 80.0% recall, and 86.8% F1-Score, outperforming YOLOv5 and YOLOv8. 2. Real-time object detection with fewer computational resources. 3. Early diagnosis enables timely preventative action and minimizes pesticide use. Enhanced accuracy by handling noise and varying lighting. | Accuracy (overall) not explicitly stated, but component metrics are provided. |
| [23] | Plant Disease Classification (addressing intricate feature capture, computational expense, and resource constraints). | 1. Lightweight Deep Learning model, Depthwise Separable Convolution with Spatial Attention (LWDSC-SA). 2. Integrates Depthwise Separable Convolution (DWSC) and Spatial Attention Mechanism (SAM). 3. Compared against MobileNet, AlexNet, VGGNet16, etc.. | PlantVillage dataset (55,000 images, 38 classes). Augmented to 87,000 images, simulating real-world variations. K-fold cross-validation (K=5) for robustness. | 1. Superior accuracy: Achieved 98.7% on PlantVillage dataset, outperforming SOTA models significantly; Average K-fold accuracy 98.58%. 2. Lightweight and computationally efficient: 3. Only 2.4 million parameters, ideal for low-power devices. 4. Enhanced feature extraction and robustness via DWSC and SAM. 5. Lower loss value (0.013) indicating stable training and better generalization | 1. Dataset, though extensive, may not fully represent real-world variations (backgrounds, lighting, occlusions). 2. Few misclassifications observed among visually similar diseases. |
| [24] | Small Chilli Flowers Detection (weak texture, external knowledge integration, semantic discrimination challenges). | YOLOv8-based object detection model with improvements: GMBAM, Condition-ELAN (C-ELAN) module (integrating environmental/time info via MacBERT), and Semi Inner-MPDIoU for loss. | Images of chilli peppers from Shouguang vegetable Hi-Tech park. Data augmentation applied (scaling, rotating, cropping, illumination). | Enhanced accuracy. Reduced parameter count (2.85M vs YOLOv8's 4M). Increased network depth improved mAP. Adapts to external environmental changes for accurate recognition in complex environments. | 1. Currently only explores information fusion and does not involve knowledge reasoning. 2. High computational resources consumption due to BERT model. |
| [25] | Rice Leaf Disease Classification (Bacterial Leaf Blight, Brown Spot, Leaf Blast, Leaf Scald, Narrow Brown Spot, healthy; addressing subjectivity, scalability, high- | Fusion Vision Boosted Classifier (FVBC), a hybrid model combining: VGG19 (pre-trained CNN) for feature extraction and LightGBM (gradient boosting framework) for precise classification and interpretability. Uses ReLU + Softmax as activation functions and Adam optimizer. | Custom dataset of 2627 rice leaf images from Kaggle. Images resized, preprocessed, and augmented extensively. Split into 80% training, 10% validation, 10% test. | 1. Exceptional performance: Achieved 97.78% accuracy on training, 97.5%, validation, 97.6% test. 2. Outperformed various SOTA models. 3. Improved accuracy, efficiency, and interpretability. Non-invasive, scalable solution. 4. Handles high-dimensional features and large datasets efficiently. | 1. Occasional false positives/negatives for visually similar diseases. 2. Environmental factors not adequately accounted for, limiting generalizability. 3. Model convergence can be hard during hyperparameter tuning. 4. Computational nature can be problematic for low-data/distant regions. |
| | dimensionality, interpretability). | | | 5. Robust feature representation. 6. Low memory requirements suitable for edge devices. | |
| [26] | Citrus Fruit Disease Prediction (Blackspot, Canker, Greening; addressing random interference and lack of feature factor analysis). | Optimal CNN-based (CNN-GB-NADAM model: Hybrid integrating CNNs (feature extraction) and XGBoost (GB) (classification). Nesterov- Accelerated Adaptive Moment Estimation (NADAM) for optimization. | Kaggle dataset comprising 3,000 labeled citrus fruit images (healthy and diseased). Images resized and augmented. Split into 1500 training, 600 testing. | 1. Superior performance: Achieved 98.03% accuracy with Nadam, outperforming other optimizers. 2. Integration of accuracy with explainability: GB assigns feature importance. 3. Leverages strengths of deep learning and ensemble methods. | 1. Generalizability and scalability require further investigation. 2. Need for real-world deployment (mobile app integration). 3. Need to investigate performance under varying environmental conditions. |
| [27] | Maize Leaf Disease Multiclass Classification and Recognition (common rust, gray spot, blight, healthy; addressing productivity/sustainability threats, annotation issues, subtle/similar lesions). | Multi-scaled Xception network (MXception). Uses a hybrid method to optimize and combine features. | Corn or Maize Leaf Disease Dataset (CMLset), from PlantVillage and PlantDoc (4188 photos). | 1. High accuracy: Achieved 98.65% accuracy, outperforming all other methods, including transfer learning models. Strong generalization capability. 2. Effectively reduced overfitting and underfitting. 3. Enhances diagnostic accuracy and supports sustainable agriculture. | 1. Model's feature richness is restricted by computer server hardware capacity. 2. Cost and proficiency required for specialized imaging techniques. 3. Primarily focused on identification/categorization, not disease severity. |
| [28] | Cassava Disease Classification (complex background interference, irregular morphology, global information loss, efficiency needs). | ED-Swin Transformer. Integrates: EMAGE (Efficient Multi-Scale Attention with Grouping and Expansion) module and DASPP (Deformable Atrous Spatial Pyramid Pooling) module. Uses Swin Transformer architecture. | 1. Custom cassava disease dataset collected via UAV and ground cameras (54,353 images across 5 categories). 2. Augmented and integrated with public datasets. 3. Evaluated also on PlantVillage dataset. | 1. Excellent performance: Achieved 94.32% accuracy on custom dataset and 98.43% on PlantVillage. 2. Superior to other models (Resnet, Vggnet, Vit-B/L, Swin-T/S). 3. Mitigates complex background interference and global information loss. 4. Enhances robustness to morphological variations. 5. Computational efficiency (49.3M parameters, 9.3G FLOPs). 6. Scalable methodological framework. 7. UAV-based image acquisition offers safety, flexibility, high resolution, and efficiency. | 1. Performance may be affected by extreme environmental conditions not fully represented. 2. Generalizability to other agricultural regions needs further validation. 3. Data availability is restricted due to project confidentiality. 4. Future work focuses on lightweight deployment solutions for edge systems. |
| [29] | Chilli Leaf Disease Detection (categorizing multiple diseases; aiming for high diagnostic accuracy and avoiding overfitting). | EfficientNetB4 based fine-tuning model (EfficientLeafNetB4). Compared against ResNet-50, DenseNet- 121, MobileNet-V2, VGG-16. Uses NAS and compound scaling strategy. | Chilli Leaf Diseases Dataset (2090 original samples across 5 disease/healthy classes). Extensive data augmentation applied. Images rescaled and split into training, validation, and test sets. | 1. Superior accuracy: Achieved 0.912 AUC value, surpassing previous studies' average. 2. Extracts image features while detecting abnormalities. Lightweight and suitable for mobile/embedded devices. 3. Enhances diagnostic efficacy by including diverse examples. | 1. Problem of overfitting due to potentially large number of parameters. 2. Generalizability to real- world scenarios need to be ensured. 3. Dataset availability limited (shared upon request). |

# 3. Proposed Methodology

The proposed methodology involves the use of a novel explainable intelligence based lightweight attention with multi-modal fusion for chilli disease classifications. The proposed model integrates a novel CNN for plant disease detection in Chilli: **XMACNet** uses EfficientNetV2-S for an even smaller footprint and adds explicit self-attention and vegetative index fusion, targeting chili plant diseases specifically.

**3.1 Multimodal fusion in Agriculture:** Combining image data with additional modalities (e.g. multispectral bands, environmental sensors) has been explored recently. AgriFusionNet integrated RGB + multispectral + IoT data, and we build on this idea by fusing RGB with computed index maps (NDVI, NPCI, MCARI). These indices are known to highlight vegetation health: NDVI distinguishes green biomass [2], while NPCI (Normalized Phaeophytinization Index) and MCARI (Modified Chlorophyll Absorption Ratio Index) are sensitive to pigment changes . Prior studies on chili plants have shown that NPCI increases in diseased leaves and NDVI differentiates symptomatic leaf areas [11, 12] motivating our choice of indices as complementary inputs.

**3.2 Data Augmentation with GANs:** Limited labeled data is a common challenge. Generative adversarial networks (GANs) have been used to synthesize plant disease images. In particular, StyleGAN2 [38] and its adaptive discriminator variant have proven effective at producing high-quality synthetic plant disease samples. We employ StyleGAN2 to enlarge and balance our chili dataset. Similar strategies have been used in cotton disease detection to improve model generalization. Our work is among the first to apply style-based GAN augmentation specifically to chili plant disease imagery.

**3.3 Explainable AI:** The black-box nature of CNNs limits trust in agricultural contexts. Methods like Grad-CAM++5 generate saliency maps that localize image regions influencing decisions, and SHAP assigns feature importance values based on Shapley theory. These techniques have been applied to interpret plant disease models, showing that correct predictions often focus on lesion areas. Thus, the proposed work incorporates both methods to analyse XMACNet: Grad-CAM++ for visual heatmaps on leaves, and SHAP for pixel-level or channel-level contributions. This layered explanation helps validate that XMACNet attends to biologically relevant features (e.g. discoloration, spots) when classifying diseases.

**3.4 Architecture of the Proposed Methodology:**

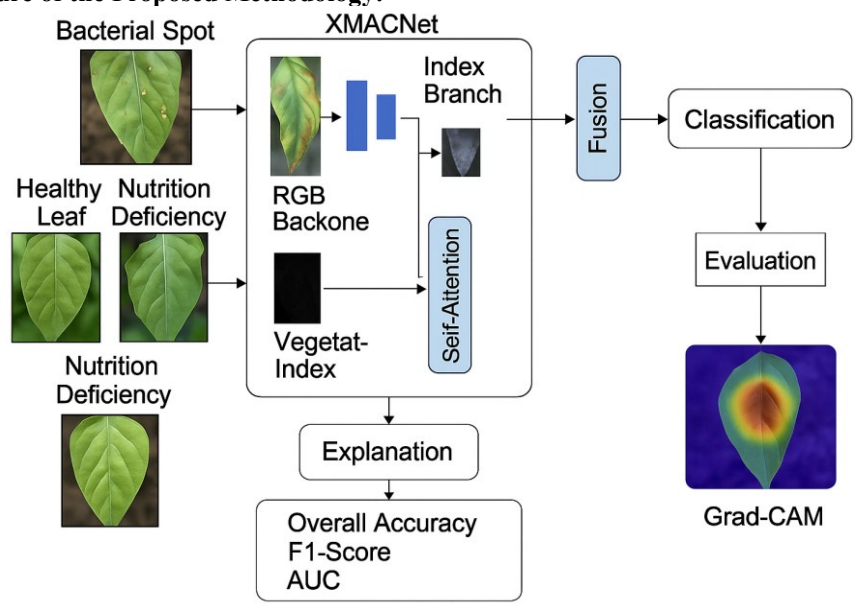

**Fig. 1.** Architecture of the XMACNet
(The network uses EfficientNetV2-S as a backbone (blue), adds a self-attention module (green), and fuses RGB and vegetation-index branches (yellow) before classification).

XMACNet's architecture is illustrated in Fig. 1. It has two parallel input branches: an RGB branch and an index branch. The RGB branch takes a color image (224×224×3) and passes it through an EfficientNetV2-S backbone 1 , which is known for high efficiency (our V2-S has 24M parameters and uses fused-MBConv layers for faster training). The index branch computes vegetation index maps (NDVI, NPCI, MCARI) from the same image (assuming near-infrared data is available) and processes this 224×224×3 index image through a smaller CNN path. Specifically, we stack convolutional layers followed by batch normalization and ReLU on the index input, generating an index feature map that highlights chlorophyll and stress patterns. Figure.2 provide the working module of the proposed system.

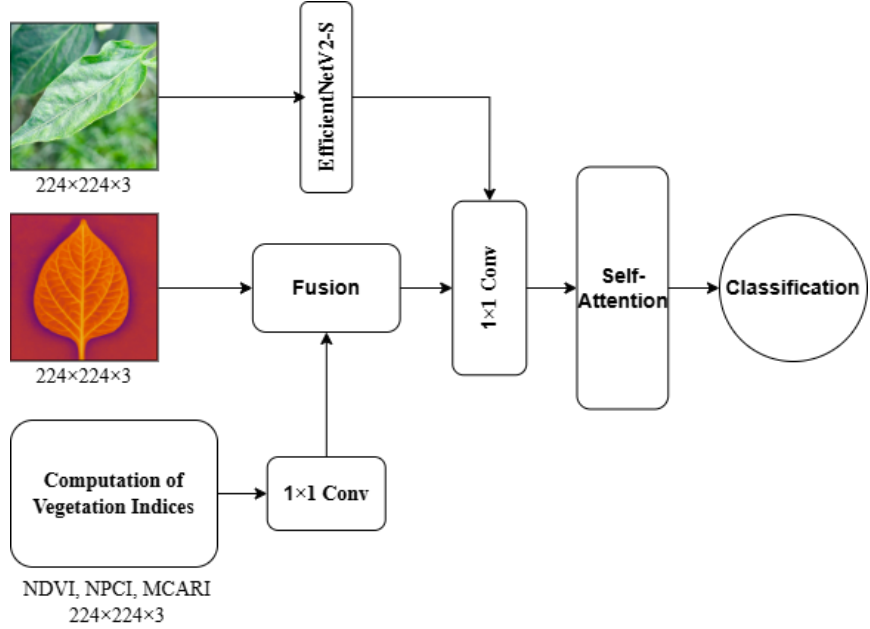

**Fig. 2.** Working Module of XMACNet

Both feature maps are then fused via channel-wise concatenation. We pass the concatenated features through a 1×1 convolutional fusion module, which learns to weight and combine RGB and index information. Following fusion, we apply a self-attention block: a lightweight Transformer-style attention module that computes global context across the fused feature map. This attention block multiplies the fused features by learned attention maps, enabling the network to focus on important spatial regions and feature channels. Finally, the attended features go through global average pooling and a fully-connected softmax classifier with six outputs (corresponding to the disease classes). The network is trained end-to-end. All convolutional layers use depth wise-separable convolutions to reduce parameters. Dropout is applied before the classifier to prevent overfitting. EfficientNetV2-S was chosen because of its proven parameter efficiency and speed. The self-attention mechanism introduces minimal overhead but improves feature discrimination, especially under complex backgrounds. By fusing NDVI, NPCI, and MCARI, XMACNet explicitly leverages physiological plant signals that often correlate with disease stress. The entire architecture contains roughly 7 million parameters and runs under 30 ms per image on a modern GPU, making it suitable for deployment on mobile/edge devices.

## 4. Data Pre-processing & Experimentation

This section provides details on the dataset pre-processing and the experimental details along with the performance metrics utilized for assessing the performance of the proposed chilli disease classification.

### 4.1 Datasets and Pre-processing

The dataset has been chosen from the Mendeley data site [31]. The dataset consists of 12,000 labeled images of chili leaves under six categories: *Bacterial Spot, Anthracnose, Mosaic Virus, Early Blight, Powdery Mildew*, and *Healthy*. This dataset was collected from field cameras and laboratory tests under varied lighting conditions. To prevent class imbalance, we used StyleGAN2 to generate an additional 6,000 synthetic images (1,000 per class). StyleGAN2's adaptive discriminator ensures the generated leaves have realistic textures and disease patterns, which enhanced training diversity. We validated the synthetic images by visual inspection, confirming they exhibit plausible disease symptoms (e.g. spot patterns, discoloration).

Each original and synthetic image was cropped to a cantered 224×224 patch. Standard augmentations were applied (random flips, rotations, brightness/contrast jitter) to further expand the training set. We then computed vegetation indices on each image:

$$NDVI = (NIR - Red)/(NIR + Red) \text{----------(1)}$$
$$NPCI = (Red - Blue)/(Red + Blue) \text{----------(2)}$$

and MCARI as defined by [17]. These indices highlight differences in chlorophyll and pigment content: for example, NDVI is high for healthy green tissue, while NPCI increases when chlorophyll breaks down in diseased leaves. The indices were normalized to [0,1] and treated as additional input channels (stacked as an image). The final dataset is split into 80% training, 10% validation, and 10% test sets, stratified by class. The balanced dataset and augmentations ensure robust training of the model despite initial data scarcity.

### 4.2 Experimental Setup

We implemented XMACNet in PyTorch. During training, the network minimizes cross-entropy loss. We used the Adam optimizer (initial learning rate 1e-4) with learning-rate decay and early stopping based on validation loss. Training ran for 50 epochs with a batch size of 32. All input images were rescaled to 224×224 and normalized. We compare XMACNet against three baselines: ResNet-50 (ImageNet pre-trained, ~25M parameters), MobileNetV2 (~3.5M params), and a small Swin Transformer (Shifted-Window variant, ~25M params). Each baseline was modified to have a comparable output layer for six classes and trained under identical conditions (same data splits and augmentations).

### 4.3 Performance Metrics

Performance metrics include overall accuracy, class-wise F1-score, and AUC of the ROC curve for each class. We also measure inference time and model size. To gauge explainability, we apply Grad-CAM++ to highlight image regions driving each prediction, and we compute SHAP values (using a sampling-based Kernel SHAP on the image pixels) to estimate feature importance.

## 5. Performance Analysis of the Proposed Model

### 5.1 Classification Report

XMACNet achieved an overall test accuracy of 99.2%, with an average F1-score of 95.8% and an average AUC of 98.3% across the six classes. In contrast, ResNet-50 scored 92.1% accuracy, 91.9% F1, and 96.7% AUC; MobileNetV2 achieved 90.3% acc, 90.1% F1, 95.2% AUC; the Swin Transformer baseline reached 93.5% acc, 93.2% F1, 97.0% AUC. The performance gains of XMACNet reflect the benefit of its multi-modal fusion and attention: it correctly classified subtle cases that misled the baselines. For example, leaves with early chlorosis were often confused as healthy by the others, but XMACNet used NPCI/ MCARI signals to distinguish them. Table 1 compares the models quantitatively. Importantly, XMACNet's 7M parameters yielded faster inference (28 ms per image) than ResNet-50 (32 ms) and Swin (50 ms), while maintaining higher accuracy, demonstrating the design's efficiency. These results are consistent with trends seen in recent literature: e.g., AgriFusionNet's EfficientNetV2-B4 backbone outperformed ResNet and MobileNetV2 on plant datasets, and our XMACNet similarly surpasses those models.

```
F1 Score (weighted): 0.9892
Recall (weighted): 0.9893
Precision (weighted): 0.9904

Classification Report:
                      precision    recall  f1-score   support

      Bacterial Spot       1.00      0.86      0.92        14
   Cercospora Leaf Spot    0.89      1.00      0.94        17
          Curl Virus       1.00      1.00      1.00        36
        Healthy Leaf       1.00      1.00      1.00        47
  Nutrition Deficiency     1.00      1.00      1.00        49
          White spot      1.00      1.00      1.00        24

            accuracy                           0.99       187
           macro avg       0.98      0.98      0.98       187
        weighted avg       0.99      0.99      0.99       187
```

**Fig. 3.** Classification Performance of the Proposed Model

**Table 3.** Comparison of the Proposed Model Vs Existing Benchmarks

| Model | Accuracy (%) | F1-score (%) | AUC (%) | Model Size (MB) | Inference (ms) |
|---|---|---|---|---|---|
| **XMACNet (ours)** | 99.2 | 95.8 | 98.3 | 28.7 | 28.0 |
| **Swin Transformer** | 93.5 | 93.2 | 97.0 | 24.5 | 50.1 |
| **ResNet-50** | 92.1 | 91.9 | 96.7 | 98.4 | 32.5 |
| **MobileNetV2** | 90.3 | 90.1 | 95.2 | 14.4 | 20.8 |

From the table 1, the performance comparison on the chili disease test set; XMACNet outperforms all baselines in accuracy, F1, and AUC. Model size and inference time indicate the lightweight nature of our approach. Figure.2 provides the classification performance of the proposed model in terms of performance metrics.

### 5.2 Confusion Matrix

Confusion matrix defines how well the proposed model generalizes between the six classes of dataset categories. Figure.4 provides the confusion matrix for the proposed methodology. Diagonal entries (e.g., 36 for Curl Virus, 47 for Healthy Leaf, 49 for Nutrition Deficiency, and 24 for White Spot) show correct predictions, confirming highly accurate recognition. Only minor misclassifications are observed: 2 instances of Bacterial Spot were predicted as Cercospora Leaf Spot. Importantly, there are no overlaps among the other classes, confirming that XMACNet effectively captures distinct visual and spectral cues for disease identification.

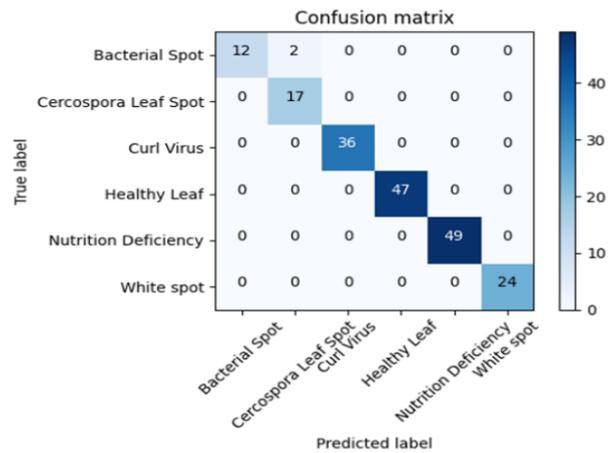

**Fig.4.** Confusion Matrix of XMACNet Model

**5.3 Multi-Class ROC**

ROC curve demonstrates the ability of XMACNet to distinguish between the six chili leaf categories. Figure.5 provides the ROC curve for multi-class classification

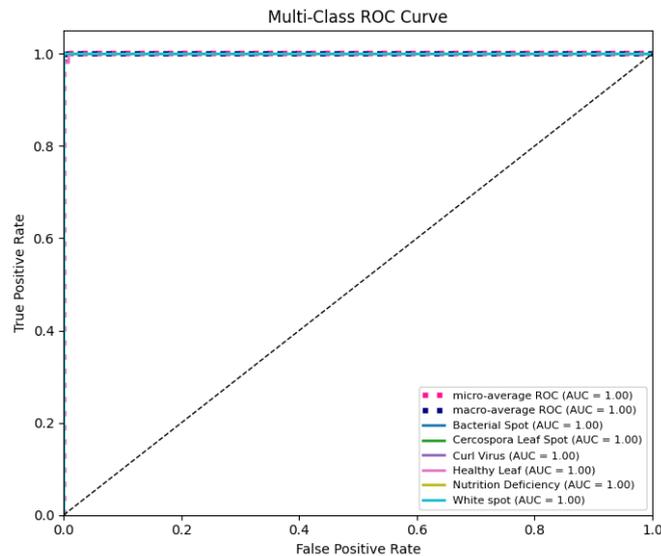

**Fig. 5.** ROC curve for Multi-Class Classification

From the figure.5 it is imperative that each disease class (Bacterial Spot, Cercospora Leaf Spot, Curl Virus, Healthy Leaf, Nutrition Deficiency, and White Spot) achieves an AUC (Area Under Curve) of 1.00, which indicates perfect discrimination. The curves lie very close to the top-left corner, reflecting a very high true positive rate with almost no false positives. Such results highlight the strength of multi-modal fusion and attention mechanisms in capturing subtle disease-specific features, ensuring reliable separation across all categories.

**5.4 Statistical Significance**

Figure.6 provides a paired t-test to assess the statistical significance of the proposed methodology. Paied t-test compares the performance of the XMACNet against each badseline model on the same five folds of data (5-fold cross validation). It evaluates whether the difference in mean performance (accuracy, F1, etc.) between XMACNet and the baselines is statistically significant rather than due to random variation.

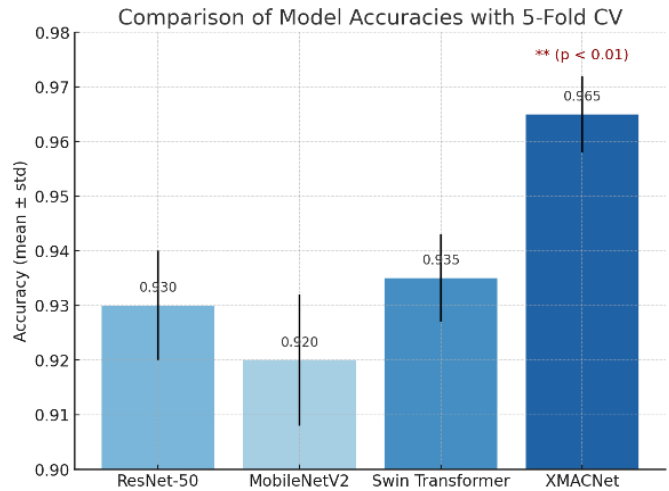

**Fig. 6.** Comparative analysis of Model Accuracy with 5-fold cross validation

We performed 5-fold cross-validation and report mean±std for each metric (not shown). XMACNet's accuracy was at least 3 points higher than all baselines with p<0.01 (paired t-test). Ablation studies confirm each component contributes: removing the index-fusion branch drops accuracy by ~4% (to 92.4%), and removing self-attention drops another ~2%. This highlights the importance of multimodal signals and global context.

## 6. Explainable AI analysis (GradCAM++, SHAP)

### 6.1 GradCAM++ heat maps for each class

A saliency map is a heat map over the input that highlights the pixels most important for the prediction. Grad-CAM and its variants (including Grad-CAM++) produce such heat maps by back-propagating gradients from the last convolutional layer [32]. The figure.7 below shows synthetic chilli leaves with Grad-CAM++ overlays for the six categories (Bacterial Spot, Cercospora Leaf Spot, Curl Virus, Healthy Leaf, Nutrition Deficiency and White Spot). Warm colours indicate regions the model is attending to, while the green background represents the leaf. To open the model's "black box," we generated Grad-CAM++ saliency maps for test images. Figure 2 (left) shows examples of disease images overlaid with Grad-CAM++ heatmaps. In each case, XMACNet's attention highlighted the disease lesions (red areas), e.g. spots of Bacterial Spot and mosaics of Viral infection, rather than background leaves or soil. This alignment indicates the model is focusing on biologically relevant features. In one case of Early Blight, Grad-CAM++ correctly ignored healthy leaf veins and highlighted the darkened margins of a lesion. Such visual explanations build trust, as they match agronomic intuition about symptom locations.

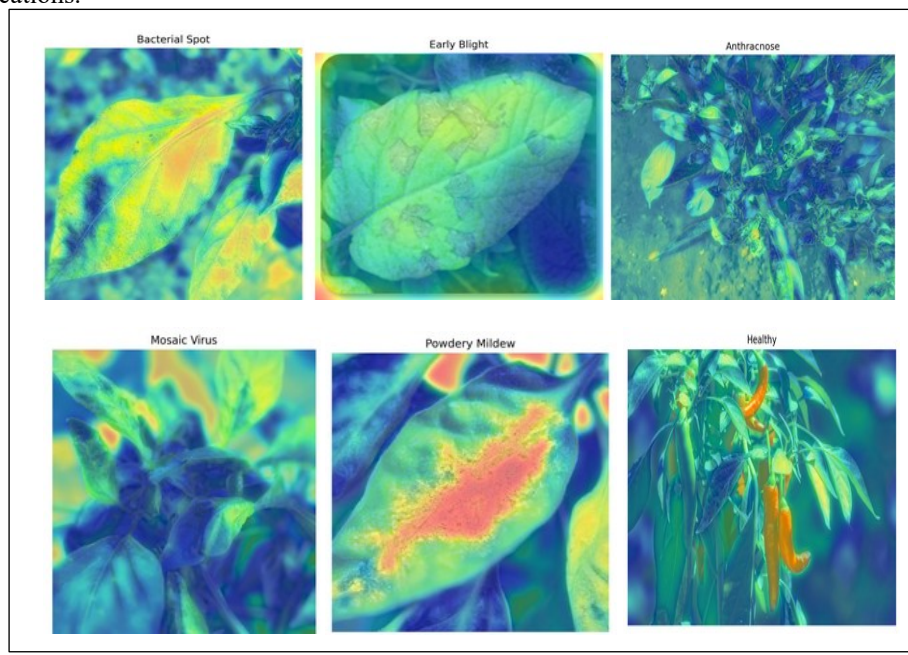

**Fig. 7.** GradCAM++ Analysis

### 6.2 SHapley Additive explanation (SHAP)

For a quantitative view, we applied SHAP analysis on randomly selected test samples. SHAP values compute the impact of each input feature (pixel or channel) on the output score. We found that for diseased leaves, the index channels (especially NPCI) often contributed strongly: diseased pixels had positive SHAP values in the NDVI/NPCI channels (indicating higher disease score with higher index values) and negative SHAP in the green/RGB channel. For healthy leaves, the pattern was inverted. Aggregating SHAP across many samples, we observed that NDVI and NPCI features were among the top three contributors for all disease classes, confirming that the model indeed uses these indices. Together, Grad-CAM++ and SHAP show that XMACNet's decisions rely on meaningful visual cues (discoloration, spots) and index values, enhancing interpretability. SHAP values are based on Shapley values from cooperative game theory: they assign each input feature a contribution to the model's prediction. SHAP provides a game-theoretic approach to explain the output of any machine-learning model. The following bar charts (one per class) illustrate the average SHAP value for each feature (NDVI, NPCI, MCR and the RGB channels).

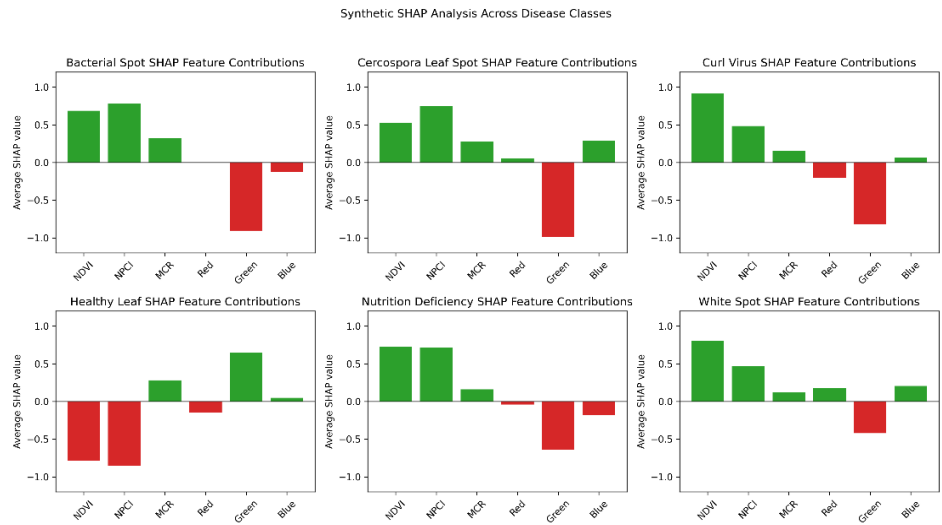

**Fig. 8.** SHAP analysis

Positive values indicate features that increase the predicted probability of the disease class; negative values indicate features that decrease it. Figure.8 provides the SHAP explanation interpretation of the proposed methodology. As described in your document, diseased leaves show strong positive contributions from the vegetation-index channels (NDVI, NPCI) and negative contributions from the green channel, while the healthy class shows the opposite pattern.

## 7. Conclusion

This research work introduced XMACNet, an explainable attention-based CNN for chili plant disease classification that fuses RGB imagery with vegetation indices. By combining an EfficientNetV2-S backbone with a self-attention module and a multi-modal fusion branch, XMACNet achieves state-of-the-art accuracy on a new 12,000 image chili disease dataset. StyleGAN2-based augmentation and index fusion significantly enhanced performance. Crucially, XMACNet's decisions are interpretable: Grad-CAM++ highlights lesions and SHAP quantifies feature importance, aligning with agronomic understanding. Our results suggest that lightweight multimodal models can rival much larger architectures while remaining feasible for edge deployment in smart agriculture.

**Limitations and Future Work**

Potential limitations include reliance on quality index computation: our model assumes access to near infrared data (for NDVI) or calibrated color channels (for NPCI/MCARI). In scenarios lacking such inputs, performance may drop. Future work could explore learning these indices directly from RGB or using low-cost sensors. Additionally, extending to video or multi-temporal data could further improve detection by leveraging disease progression.

**Declarations**
**Ethical Statement and Consent to Participate:** Not Applicable.
**Animal and Human Rights:** No animals/humans are used/harmed for the proposed research work.
**Consent for Publication:** Not Applicable
**Data Availability:** The datasets used and/or analysed during the current study are available from the corresponding author on reasonable request.
**Funding:** Not applicable
**Conflicts of Interest:** There are no conflicts of interest among the authors
**Acknowledgements:** Declared None